\documentclass[11pt]{article}
\usepackage[numbers]{natbib}
\usepackage{amsmath, amssymb, amsthm}
\usepackage{graphicx}
\usepackage{bm}
\usepackage{enumitem}
\usepackage{geometry}
\usepackage{listings}
\usepackage{xcolor}
\usepackage{hyperref}
\usepackage{mathpazo}

\lstdefinestyle{pythonstyle}{
  language=Python,
  basicstyle=\ttfamily\small,
  keywordstyle=\color{blue},
  commentstyle=\color{gray},
  stringstyle=\color{orange},
  showstringspaces=false,
  columns=fullflexible,
  keepspaces=true,
  frame=single,
  rulecolor=\color{black},
  tabsize=4,
  breaklines=true,
  breakatwhitespace=true
}

\title{\huge Prioritized Replay for RL Post-training}
\author{Mehdi Fatemi \\ \texttt{mehdi.fatemi@ieee.org}}
\date{}

\begin{document}
\maketitle

\begin{abstract}
We introduce a problem-level prioritization framework for RL post-training of large language models. Building on insights from prioritized replay in deep RL, as well as prior observations that rollouts with intermediate success rates tend to produce stronger learning signals under methods such as GRPO, our approach selects problems according to a simple, model-driven priority score derived from empirical success statistics. In contrast to conventional curriculum strategies that emphasize easier tasks early in training, the resulting schedule naturally focuses training on problems that are neither consistently solved nor consistently failed, while deprioritizing those that contribute little gradient information. The method yields a continuously adapting and automatic prioritization process that requires no predefined difficulty tiers, auxiliary predictors, or external labels. We further introduce lightweight mechanisms for practical deployment, including heap-based prioritized sampling and periodic retesting of solved and unsolved problems to mitigate starvation and forgetting. Overall, the approach offers a principled and scalable alternative to manually designed curricula while aligning data selection directly with the dynamics of GRPO-based post-training.
\end{abstract}

\section{Introduction}

In many reinforcement learning (RL) settings, the efficiency of the training process depends critically on how learning examples are selected. A well-established result from the deep RL literature is that prioritized sampling can substantially improve learning speed and final performance. The canonical example is Prioritized Experience Replay (PER) by Schaul et al.\ \cite{schaul2015per}, which replaces uniform sampling from the replay buffer by sampling transitions in proportion to a priority measure derived from temporal-difference (TD) error. PER yielded strong empirical gains on the Atari 2600 suite. When integrated into DQN, the median normalized score increased from $48\%$ under uniform sampling to $106\%$ under prioritized sampling, with improvements in $41$ out of $49$ games. These benefits persisted even within the more advanced Rainbow agent \cite{hessel2018rainbow}, where PER remained one of the most influential components in the ablation analysis. Although recent work shows that uniform replay may outperform PER in some settings \cite{zhang2024investigating}, the broader empirical trend remains that prioritization often provides a significant advantage. Moreover, uniform sampling can be combined with PER to inject beneficial exploration, an idea closely related to several extensions considered in this work.

We study RL post-training for large language models (LLMs), where the environment is a collection of reasoning problems or task simulators that produce complete rollouts for the agent. This framework differs from classical value-based RL, since we do not maintain a replay buffer of state-action transitions or rely on TD errors. Instead, the environment's main role is to select which problems or tasks the model trains on at each training step. Because of this structural shift, transition-level prioritization is no longer appropriate and priority must be defined at the problem level. The prioritization scheme we propose satisfies this need and is compatible with GRPO \cite{grpo_shao2024}, PPO \cite{ppo_schulman2017}, or any other rollout-based method, since it governs the scheduling of problems rather than the form of the learning update.

Prioritized replay offers a systematic and model-driven alternative to handcrafted curricula, replacing manual curriculum design with a dynamic and automatic process that adapts directly from data and from the capability of the model at each point in time. 
Recent work has explored improving the efficiency of GRPO through difficulty-targeted sampling and rollout reuse, including approaches that estimate problem difficulty using auxiliary models and prioritize problems whose estimated success probability is close to $50\%$ \cite{sun2025}. Our work is complementary to this line of research. Rather than learning a separate difficulty estimator or modifying the training loss, we study a non-parametric formulation in which the priority score arises directly from the GRPO group-advantage structure and is implemented as a lightweight problem-level replay scheduler. Alongside the core prioritization scheme, we introduce several practical mechanisms, including the retention and periodic reevaluation of fully solved and currently unsolvable problems and the deliberate use of exploration. These additions help prevent starvation and over-concentration effects and help ensure that prioritization remains effective throughout training.

Finally, because of compute constraints, our empirical evaluation is intentionally limited and should be interpreted as a proof of concept rather than a comprehensive experimental study. A systematic examination of configuration choices, particularly learning rates and the management of exploration, goes beyond our current budget, even though these factors are important for the effective use of prioritized replay in large-scale post-training settings.

\section{Core Method}

The goal is to identify, at each training iteration, the subset of problems that are expected to provide the strongest learning signal under the GRPO update rule. Motivated by prior analyses highlighting the importance of problems of intermediate difficulty for GRPO \cite{sun2025}, we examine the problem-level advantage structure and derive a scalar quantity that measures the informativeness of a problem as a function of the model's empirical success probability.

Consider a fixed problem and assume that we generate $N$ independent responses. Each response receives a binary reward $r_i \in \{0,1\}$, where $r_i = 1$ indicates a correct solution. Let
\[
p = \frac{1}{N}\sum_{i=1}^N r_i
\]
denote the empirical success rate for this problem in the group of size $N$. GRPO defines the group advantage for response $i$ as
\[
A_i = r_i - p.
\]
The policy gradient is then linearly weighted by $A_i$. A basic property of this definition is that advantages sum to zero,
\[
\sum_{i=1}^N A_i
   = \sum_{i=1}^N (r_i - p)
   = \sum_{i=1}^N r_i - Np
   = Np - Np
   = 0.
\]
Thus the GRPO update extracts information purely from the variation within the set of rewards generated for a single problem. If all $r_i$ are identical (if the model always succeeds or always fails on this problem), then each $A_i$ is zero and the problem yields no gradient signal. Conversely, if the model is correct on some responses and incorrect on others, the advantages take nonzero values and a useful learning signal is produced. Of note, dividing $A_i$ by the standard deviation preserves its zero-sum property and merely scales the resultant gradient magnitude \cite{fatemi2025pgbaseline}.

To quantify this variation, we consider the mean squared advantage
\[
\frac{1}{N}\sum_{i=1}^N A_i^2
  = \frac{1}{N}\left(Np(1-p)^2 + N(1-p)p^2\right)
  = p(1-p).
\]
This quantity is the variance of a Bernoulli random variable with mean $p$ and reaches its maximum at $p = \tfrac{1}{2}$. It vanishes at $p=0$ and $p=1$, matching the intuition that a problem the model always solves or always fails provides no gradient signal for GRPO. We therefore define the priority score of a problem as
\[
\omega = p(1-p).
\]
Problems with success rates near one-half receive high priority, whereas those with extreme success rates receive low priority. Because $\omega\in[0,1/4]$, the priority score is naturally bounded and comparable across problems. In contrast to approaches that infer difficulty using auxiliary predictive models or similarity-based estimators (e.g., see \cite{sun2025}), the priority used here is computed purely from empirical success statistics gathered during training, making the mechanism fully non-parametric and inexpensive to deploy.

At the beginning of training, no empirical estimate of $p$ is available for any problem. Pre-estimating $p$ through multiple rollouts is computationally wasteful for large datasets, and the estimate can instead be yielded online as training progresses. To guarantee that every problem is sampled at least once, all priority scores may be initialized to $+\infty$. After a problem has been sampled, its priority score is updated to the finite value $\omega = p(1-p)$ computed from the observed responses. To reduce noise across iterations, $p$ may optionally be maintained using an exponential moving average. As a more efficient alternative, priority scores can be initialized to any value below $0.25$, the maximum of $p(1-p)$. This initialization allows the algorithm to immediately exploit genuinely high-priority problems as soon as they are identified, without requiring a complete sweep of the dataset. However, this choice no longer guarantees that every problem is forcefully visited at least once. Note also that choosing an initialization value below $0.25$ controls the trade-off between uniform first-time sampling and early exploitation of problems whose priorities exceed that threshold. For example, for a group size of $N=8$, initializing priorities at $0.2$ allows problems with $3$, $4$, or $5$ correct responses to take precedence, while previously unseen problems will not be sampled until later model updates reduce the priorities of the currently dominant problems below $0.2$ (corresponding to correctness levels outside the $3$--$5$ range). Additional exploration strategies are discussed in Section~\ref{sec:considerations}.

\section{Computational Deployment}

To efficiently obtain, at each iteration, the $C$ problems with the largest priority scores, we employ a binary max-heap, following \cite{schaul2015per}. A binary heap is an array-based representation of a complete binary tree satisfying the heap property that for each index $i$, the parent has priority score no smaller than its children. If indices start at zero, the children of position $i$ reside at positions $2i+1$ and $2i+2$. The maximal-priority element is always stored at index $0$. The operations \texttt{extract\_max} and \texttt{insert} both require $O(\log M)$ time for a heap of size $M$, and are implemented using the standard sift-down and sift-up procedures, which move an element downward or upward until the heap property is restored.
The basic training iteration proceeds as follows:
\begin{enumerate}
    \item Use \texttt{extract\_max} repeatedly to obtain the $C$ problems with the highest current priority scores. These are removed from the heap.
    \item For each extracted problem, generate $N$ responses, compute their binary rewards, and apply the GRPO update to the model.
    \item For each such problem, compute a new priority score $\omega = p(1-p)$, where $p$ is the empirical (or exponentially averaged) success rate.
    \item Reinsert each problem into the heap using \texttt{insert} together with its newly computed priority score.
\end{enumerate}

Because a problem's priority score is always updated immediately after extraction (never while the problem is still located within the heap) no random deletion or in-place priority modification is required. Consequently, the heap implementation does not need to maintain auxiliary index maps from problem identifiers to heap positions, and its memory footprint consists only of the priority score and the heap arrays. We later discuss exploration, which essentially requires random deletion. Nevertheless, the additional memory need remains negligible even for large $M$.

\section{Important Considerations} \label{sec:considerations}

\subsection{Forgetting and Starvation}

Two additional refinements of the priority mechanism address both fully solved and currently unsolvable problems. A problem for which the model achieves $p=1$ consistently yields $\omega=0$ and therefore provides no learning signal. Rather than reinserting such problems into the priority heap, it is useful to maintain a separate pool of fully solved problems. During normal training these problems are ignored, but at a low frequency one may draw a small random batch from this pool and regenerate rollouts. If a problem in this batch remains fully solved, no gradient update is performed; if the model begins to fail on it, then its new empirical success rate determines a positive priority score, and the problem is reinserted into the main heap. This mechanism helps guard against regression on previously mastered material and can reduce the risk of forgetting.

A symmetric consideration applies to problems that currently appear unsolvable. Such problems exhibit $p=0$ and hence also $\omega=0$, which places them permanently at the bottom of the max-heap and effectively prevents them from being revisited. To avoid this starvation effect, these problems may similarly be removed from the heap and stored in an unsolved pool. Periodically, a random subset of this pool can be tested again. If a problem still yields $p=0$, it remains in the unsolved pool; if the model succeeds on at least some responses, that problem receives a positive priority score and is reintroduced into the heap, making it eligible for regular prioritized training. In practice, small tolerance bands such as $p \le \varepsilon$ or $p \ge 1-\varepsilon$ can be used instead of exact zero or one to account for estimation noise.

Both the solved and unsolved pools can be efficiently organized as independent min-heaps keyed by the timestamp of their most recent evaluation. This setup ensures that problems that have been checked most recently are selected again only after all others have been revisited.

\subsection{Exploration}

A further modification introduces controlled exploration directly within the priority heap. Instead of always selecting the problems with the highest priority scores, one may occasionally sample a problem uniformly at random from the heap and temporarily remove it. This requires support for arbitrary deletion, which in turn increases the memory footprint by maintaining a mapping between problem identifiers and their heap positions. Such exploration can prevent the training loop from focusing too narrowly on a small subset of problems whose priority scores fluctuate around intermediate values, and may help the model discover informative problems that would otherwise remain at consistently low priority.

These refinements may be necessary because the model is constantly updated and improved. As a result, old priority scores may no longer be accurate, particularly for problems which have lower priority and hence have not been tried recently.

Alternatively, one may replace deterministic max-priority extraction with a stochastic scheme based on a sum-tree structure that supports priority-proportional sampling. In this setup, the probability of selecting a problem is proportional to its current priority score $\omega$, providing a balance between exploitation and exploration. Additional parameters may also be introduced to tune this trade-off. Whether deterministic extraction of the highest-priority problems or stochastic sampling proportional to their priority yields better empirical performance depends on the characteristics of the task distribution and remains an open design choice.

It is also worth emphasizing that, in practice, the dominant computational cost of the environment procedure arises from generating and evaluating model rollouts. These operations are orders of magnitude more expensive than any of the priority-management mechanisms described above. As a consequence, the choice between a deterministic max-heap, an exploration-enhanced heap with arbitrary deletions, or a sum-tree supporting priority-proportional sampling should be driven primarily by empirical training dynamics rather than computational considerations. All of these data structures are highly efficient, and their overhead is negligible relative to the cost of producing rollouts. Therefore, any variant that improves training behavior can be adopted without practical concern about its impact on runtime.

\subsection{Learning Rate}

The use of prioritized replay directly amplifies the magnitude of the learning signal. Because prioritized sampling favors problems with high advantage variance, the expected per-update advantage under prioritized GRPO is larger than it would be under uniform sampling. Since the policy-gradient loss scales linearly with advantage, this amplification translates into larger effective gradients during optimization. A practical implication is that learning rates tuned for standard GRPO may become overly aggressive when used with prioritized replay. Without adjustment, the interaction between higher gradient magnitudes and unchanged hyperparameters can increase training instability. We have observed this effect when applying off-the-shelf configurations commonly used in frameworks such as \texttt{verl} \cite{sheng2024hybridflow}.
In practice, it may therefore be necessary to reduce the base learning rate or apply adaptive learning-rate schedules when enabling prioritized sampling. More generally, hyperparameters that implicitly depend on gradient scale, such as clipping ranges, entropy coefficients, or EMA smoothing constants, may also require retuning to ensure stable training dynamics under prioritized replay.

\subsection{Concise Reasoning}

Concise reasoning has recently become a central objective in the development of reasoning-oriented language models \cite{concise-reasoning2025}. The term refers to achieving high problem-solving performance while producing substantially fewer tokens, which in turn improves resource usage, cost, latency, and the interpretability of chain-of-thought traces, and can even enhance reasoning quality by reducing redundancy and irrelevant segments. As shown in \cite{concise-reasoning2025}, continued training on correct responses causes the optimization dynamics to favor shorter and more direct solutions. In contrast, reinforcement on incorrect responses tends to produce unnecessarily verbose outputs. Somewhat counterintuitively, this effect arises solely from the structure of the loss and the advantage sign, independent of any explicit incentive toward brevity or even toward exploration.

Motivated by this phenomenon, it can be desirable to introduce a slight bias toward problems on which the model already performs relatively well. In the core method described above, the priority score $\omega = p(1-p)$ is perfectly symmetric around $p = 1/2$. For example, in a group of $N=8$, a problem with $k=6$ correct responses ($p=0.75$) receives the same priority score as a problem with $k=2$ correct responses ($p=0.25$), since both yield $\omega = \tfrac{6}{8}\cdot\tfrac{2}{8} = 0.1875$. However, if concise reasoning is desired as a secondary objective, it may be preferable to prioritize the former problem to reinforce rollouts with a higher rate of correct answers.

There are several ways to introduce this bias without fundamentally altering the prioritization mechanism. One simple approach is to add a small positive constant to the priority score whenever $k \geq N/2$. This preserves the maximum at $p=1/2$ while breaking the symmetry between $p$ and $1-p$ in a controlled manner. The constant can be set sufficiently small so that it merely breaks ties between symmetric cases of $k$ and $N-k$ without affecting the ranking of other pairs. This lightweight adjustment nudges the prioritized replay process toward promoting concise reasoning while retaining the core benefits of the variance-based priority measure.

\section{Experiments}

\begin{figure}[t]
\centering
\includegraphics[trim=0 0.12in 0 0.1in, clip]{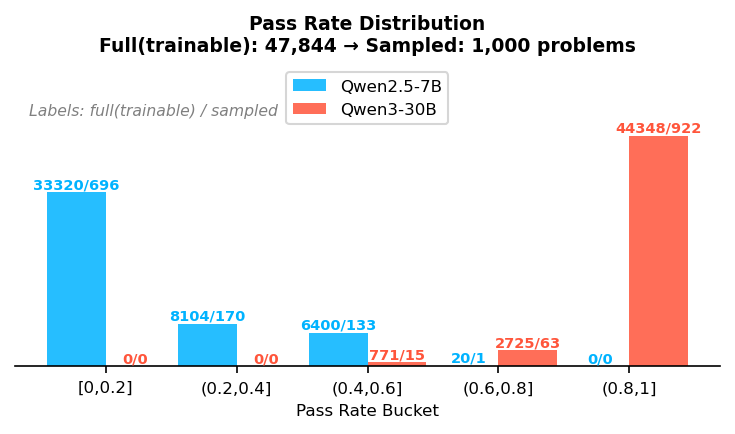}
\caption{Success-rate distributions of the training data from the trainable subset of the GURU dataset \cite{cheng2025guru} and a stratified 1,000-sample sub-subset, shown from the perspective of both models.}
\label{fig:train-dist}
\end{figure}

This section provides a proof-of-concept evaluation under a constrained compute budget. The goal is to assess the behavior of prioritized replay in a realistic but limited training setting rather than to perform large-scale hyperparameter optimization. Effective deployment in larger training regimes may require careful tuning of learning rates, exploration behavior, and related choices in order to obtain the full benefit of prioritization.

We fine-tune Qwen2.5-7B \cite{qwen25} using GRPO on a stratified subset of 1000 problems from the GURU dataset \cite{cheng2025guru}. GURU contains 54K math problems with different levels of difficulty and reports two success rates ($p_{guru}$) for each problem, estimated using 64 samples from two models: Qwen2.5-7B and the stronger Qwen3-30B \cite{qwen3}. The dataset additionally applies several filtering steps, including the removal of duplicates and problems that achieve $p_{guru}=0.0$ on Qwen3-30B. From the original set of 54,404 math problems, we select a smaller subset of 47,844 problems with $p_{guru}>0.5$ on Qwen3-30B. This selection yields problems that are reasonably learnable under our smaller-scale training regime. Despite this filtering, the subset still contains a large fraction of fully or nearly unsolvable problems for Qwen2.5-7B (over $60\%$; see Figure \ref{fig:train-dist} for the full distribution). Finally, for our experiments, we select a random subset of 1000 problems, stratified to match the difficulty distribution of the full dataset. \textbf{We do not use $p_{guru}$ or other labels for training}.

For each training step, we sample a batch of 4 problems and generate 8 rollouts per problem at temperature 0.7, yielding 32 response-reward pairs per step. We use $A_i = r_i - p$ without standard-deviation normalization, where $p$ is the group mean reward. The reward function assigns binary scores based on answer correctness, using the Qwen Math Eval Toolkit \cite{yang2024qwen2} for robust symbolic equivalence checking. For the codebase, we forked from \texttt{verl}  \cite{sheng2024hybridflow} and prioritized sampling.

Success rates are tracked using exponential moving average with $\alpha = 0.8$. Problems are prioritized by $\omega = \bar{p}(1-\bar{p}) + \epsilon$, where $\bar{p}$ is the EMA success rate and $\epsilon = 10^{-4}$ is a small bias when $p\ge 0.5$. All problems are initialized at $\omega=0.2$. As training advances, problems with $p = 1$ (fully solved) or $p = 0$ (fully unsolved) are moved to separate pools and periodically retested (every 10 steps, 1 from the solved pool and 3 from the unsolved pool). Exploration is performed at the batch level: 12.5\% of training batches are sampled uniformly at random from the main heap rather than by priority. The baseline uses identical hyperparameters but with uniform random sampling instead of prioritized sampling.

Models are evaluated every 25 training steps on the MATH-500 benchmark (500 problems) \cite{hendrycksmath2021} and AIME 2024 (30 problems) \cite{aime24}, using 4 rollouts per problem at temperature 0.7. We use AdamW with learning rate $10^{-6}$, no weight decay, no learning rate schedule, and no KL penalty. PPO clipping ratio is 0.2 with gradient clipping at 1.0. Training runs for 250 steps on 8$\times$B200 GPUs. Note that with 250 training steps and 4 problems per step, the baseline setting covers the entire dataset once, whereas prioritized sampling will only expose the model to a subset of it.

\begin{figure}[t]
\centering
\includegraphics[width=5.5in, trim=0 0.1in 0 0.1in, clip]{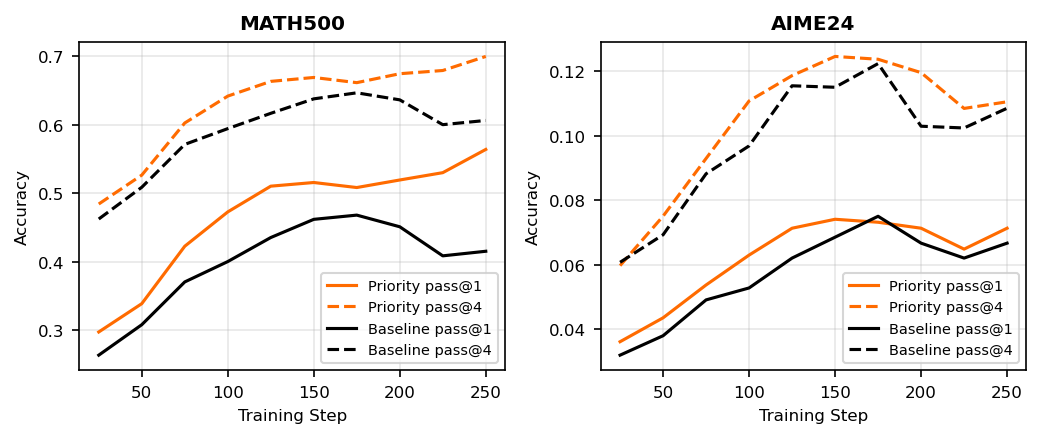}
\caption{Evaluation on MATH500 and AIME 2024 datasets during training steps.}
\label{fig:eval-acc}
\end{figure}

\begin{figure}[t]
\centering
\includegraphics[width=6in, trim=0 0.1in 0 0.1in, clip]{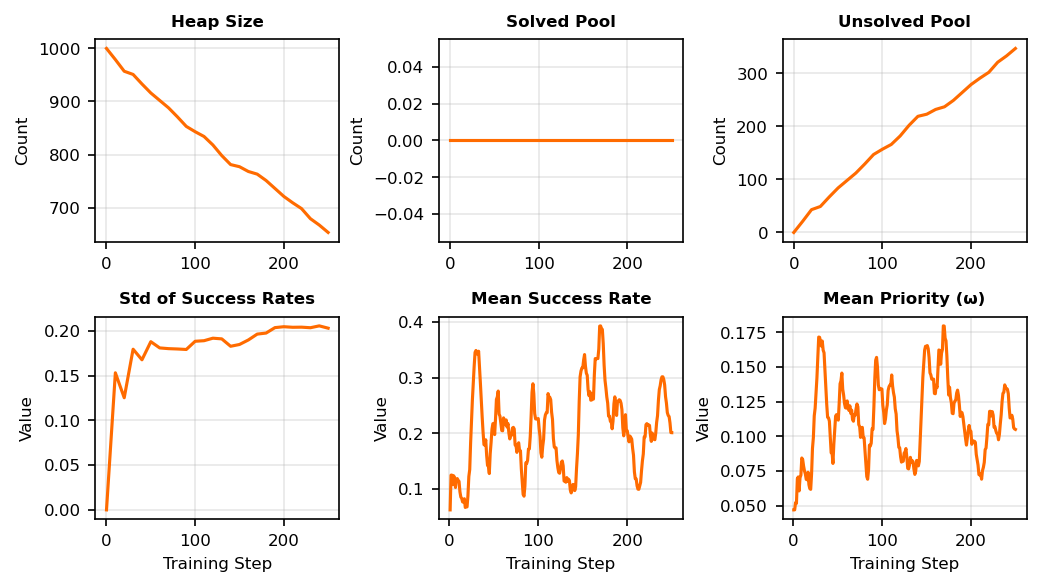}
\caption{Dynamics of the prioritized sampling during the training process.}
\label{fig:priority-dynamics}
\end{figure}

\subsection*{Results}

Figure \ref{fig:eval-acc} presents \textbf{pass@1} and \textbf{pass@4} performance of the uniform baseline and the prioritized sampling method, with curves smoothed using a window of 3 prior values. Within the reported 250 steps, the prioritized sampler almost always outperforms uniform sampling on both benchmarks and for both pass@1 and pass@4. 

The six plots in Figure \ref{fig:priority-dynamics} provide insight into the internal dynamics of prioritized sampling during training. All curves are smoothed using a window of ten prior values to reduce noise.

\textbf{Row 1.} The heap size denotes the number of learnable problems, which are those with intermediate success rates $0 < p < 1$ that produce meaningful gradients. As training progresses, problems gradually migrate to the solved pool ($p = 1$) or the unsolved pool ($p = 0$). A healthy training trajectory is characterized by a steady flow from the heap into these pools as the model's competence boundaries become more clearly defined. In our small scale experiment with only 250 training steps, none of the problems reached full success and therefore none entered the solved pool, while many problems were progressively classified as unsolvable. We also observe that more than 600 problems remain unseen during this run, because the initialization strategy prioritizes problems with 3--5 correct answers, when such problems are available, instead of unseen problems. This sampling bias may also help explain the behavior observed beyond step 250 in our briefly extended but incomplete runs, where performance oscillates without substantial improvement. Limitted exposure to unseen data and potentially the use of a large learning rate may be contributing factors that require further investigation.

\textbf{Row 2.} The standard deviation of success rates reflects the diversity of problem difficulty as perceived by the model. High values (approximately 0.4 to 0.5) indicate a wide spread of difficulties, with some problems being easy (high $p$), some hard (low $p$), and others in between. This situation is generally desirable for prioritized sampling, because it allows the sampler to meaningfully distinguish among problems. In contrast, a low standard deviation indicates that most problems have similar success rates, which may occur when problems are uniformly easy ($p \approx 1$), uniformly hard ($p \approx 0$), or clustered around an intermediate difficulty level.

In this experiment, the std of success rates remains relatively small throughout training, which is consistent with the fact that all problems are initialized with equal success estimates and many problems remain unseen within the 250 training steps. However, the observation that the standard deviation stabilizes near 0.2 is concerning, as it suggests limited exploration of unseen data. The plot indicates that, in the last 60 steps of training, the active portion of the heap is dominated by a set of solvable problems whose priorities remain higher than the default priority assigned to unseen problems. As a result, although unseen problems are still present in the heap, they are rarely selected for training. At the same time, the 12.5\% mechanical exploration does not appear to compensate for this effect. This is likely because a substantial fraction of the dataset is truly unsolvable and small exploration tends to add additional problems to the unsolved pool rather than enriching the set of learnable problems, which is also reflected in the steadily (nearly linear) increasing number of items classified as unsolved.

\section{Discussion}

This work presents a problem-level formulation of prioritized replay for RL post-training and shows that the method can be integrated into a realistic training pipeline. In our experiments, prioritized sampling provides stronger early-stage performance than uniform sampling on both MATH500 and AIME 2024 for pass@1 and pass@4 under the same configuration that is commonly used with uniform sampling. These results should be viewed as an initial proof of concept rather than a comprehensive evaluation. Our compute budget limited the range of configurations we could explore and constrained training to a relatively small number of steps. We did not perform extensive hyperparameter sweeps or multi-seed studies, so the reported results should not be interpreted as fully optimized.

Training runs were only briefly extended beyond the 250-step window shown in the figures. In these longer runs, the performance of prioritized sampling began to fluctuate and appeared to plateau. We did not investigate this regime in a systematic manner, but our preliminary observations suggest that late-stage behavior in prioritized sampling may be sensitive to factors such as learning rate, exploration schedule, and the evolving composition of solved and unsolved pools. With larger-scale experiments and additional tuning, it is plausible that these fluctuations could be prevented. Importantly, we interpret these findings not as evidence against prioritized sampling, but as an indication that it may require configuration choices that differ from those typically used with uniform sampling and may also call for broader experimentation with the exploration frequency. The present work focuses on introducing the conceptual framework and demonstrating its feasibility in a controlled experimental setting.

Overall, our formulation provides a simple and complementary alternative to difficulty-prediction and rollout-replay approaches such as \cite{sun2025}, offering a replay-centric mechanism that incorporates difficulty awareness directly through the GRPO group-advantage structure.

\section{Code and Reproducibility}

The code that generates our experimental results is available at this fork:

\href{https://github.com/fatemi/verl/tree/feature/priority-sampling}{\texttt{https://github.com/fatemi/verl/tree/feature/priority-sampling}}

\noindent See the \href{https://github.com/fatemi/verl/tree/feature/priority-sampling/examples/priority_sampling}{\texttt{examples/priority\_sampling}} subfolder.

The GURU dataset \cite{cheng2025guru} is publicly available through huggingface. All the details to download and stratify train data as well as the test data we used are provided in the \texttt{guru.ipynb} notebook.

\section{Related Work}

Prioritization in training data selection has long been recognized as a driver of learning efficiency in machine learning and reinforcement learning. In deep RL, Prioritized Experience Replay \cite{schaul2015per} is a well-established method that samples experience transitions in proportion to their temporal-difference error, yielding significant empirical gains. PER operates at the transition-level granularity, which makes it ill-suited for sequence models trained on long-horizon reasoning tasks. For such tasks, temporal-difference errors from individual tokens or small transitions do not always yield an informative measure of a problem's overall usefulness for learning. Summing or averaging token-level temporal-difference errors over an entire response is ad hoc and can obscure the informativeness of the example as a whole.

In contrast, our work introduces a principled problem-level prioritization scheme tailored for RL post-training of language models on reasoning tasks. Unlike PER and its variants, our method computes priority directly from model performance at the whole-problem level. It naturally emphasizes problems with intermediate difficulty, those most likely to yield strong learning gradients, and deprioritizes problems that are either always solved or always failed. This scheme is compatible with rollout-based training algorithms like GRPO and PPO and does not require maintaining a buffer of low-level transitions or computing token-wise temporal-difference errors.

A recent line of work investigates improving the efficiency of GRPO-based RL post-training through adaptive sampling and data reuse. Sun et al.~\cite{sun2025} introduce a framework that predicts problem difficulty using an auxiliary model, prioritizes problems whose estimated success probability is near $50\%$, and combines this with rollout replay to reduce data and compute requirements. Their method emphasizes difficulty prediction and replay at the trajectory level. Our approach is complementary. We study a non-parametric, replay-centric formulation in which difficulty arises directly from empirical success rates and the GRPO group-advantage structure, and we focus on the scheduling and stability properties of a lightweight heap-based mechanism with solved/unsolved pools.

Several recent works propose adaptive curricula or data selection policies in the context of language model training. For instance, \citet{chen2025selfcurriculum} introduce a self-evolving curriculum policy that treats task selection as a multi-armed bandit problem, selecting from predefined task categories based on policy advantage. Their method operates at the level of task categories, not individual problems. Similarly, \citet{wang2025dump} propose a distribution-level curriculum over data sources using UCB-based exploration, again assuming predefined task partitions. In contrast, our method works at the finest level of granularity, individual reasoning problems, which allows for more precise targeting of learning potential.

A related line of work is curriculum reinforcement learning, where training progresses from easier to harder tasks according to a predefined or learned schedule \cite{bengio2009curriculum,narvekar2020curriculum}. E2H (Easy-to-Hard) \cite{parashar2025e2h} follows this paradigm in the context of LLM RL by grouping problems into difficulty tiers and using a probabilistic scheduler that gradually shifts focus from easy to hard tasks. Empirically, E2H reports substantial gains for small LLMs across several reasoning benchmarks when compared to vanilla RL training, and its theoretical analysis within an approximate policy-iteration framework shows that, under suitable task decompositions, such curricula can reduce sample complexity relative to training directly on the hardest tasks \cite{parashar2025e2h}. However, the core curriculum assumption that easier tasks should be emphasized earlier is not directly aligned with the source of gradient signal in multi-sample advantage-based methods such as GRPO or PPO. For these algorithms, as shown in this paper, the most informative problems are those with a balanced mix of correct and incorrect responses. Moreover, this set of balanced problems changes continuously as the model is being updated. Static human-assigned difficulty labels are inherently limited in that they may be noisy or mismatched to the specific base model being trained, and they do not reflect the model's evolving competence. In contrast, our approach (1) uses a model-driven priority score computed from rollouts, (2) requires no manual labels or curriculum schedules, and (3) adapts automatically as performance on each problem changes. It can be viewed as a model-driven alternative to label-based curriculum learning.

Additionally, we implement retention mechanisms for both fully solved and currently unsolvable problems. Problems with consistently high or low performance are removed from the active training set and re-tested periodically. This guards against catastrophic forgetting and ensures that the model can revisit problems that may have become tractable. Such mechanisms are absent from many prior curriculum or prioritization methods, including E2H and SEC.

Our implementation also builds on the efficient sampling mechanisms introduced in \citet{schaul2015per}, employing binary heaps or sum-tree structures to maintain priority queues with minimal overhead. This makes the method practical and scalable.

Finally, unlike methods that require a separate policy to learn the curriculum (for example \citet{lv2025raise}), our method avoids additional training components and fits cleanly into existing RL fine-tuning pipelines.

In summary, our work distinguishes itself through the following aspects.
\begin{itemize}
\item Problem-level granularity. Unlike prior work that prioritizes task categories or data sources, we operate at the level of individual problems.
\item Practicality. No labeling, no additional policies, and no token-level temporal-difference computations, which do not exist in GRPO.
\item Built-in forgetting resistance. Re-testing mechanisms for solved and unsolved problems.
\item Dynamic adaptation. Priority is updated continuously from observed success, enabling an emergent curriculum that follows the model's competence.
\end{itemize}
These distinctions make our method both principled and immediately applicable to RL post-training of reasoning-capable language models.

\clearpage
\bibliographystyle{unsrtnat}
\bibliography{references}

@inproceedings{schaul2015per,
  title={Prioritized Experience Replay},
  author={Schaul, Tom and Quan, John and Antonoglou, Ioannis and Silver, David},
  booktitle={International Conference on Learning Representations},
  year={2016},
  url={https://arxiv.org/pdf/1511.05952.pdf}
}

@inproceedings{hessel2018rainbow,
  title={Rainbow: Combining Improvements in Deep Reinforcement Learning},
  author={Hessel, Matteo and Modayil, Joseph and van Hasselt, Hado and Schaul, Tom and Ostrovski, Georg and Dabney, Will and Horgan, Dan and Piot, Bilal and Azar, Mohammad Gheshlaghi and Silver, David},
  booktitle={AAAI Conference on Artificial Intelligence},
  year={2018},
  url={https://arxiv.org/pdf/1710.02298.pdf}
}

@article{zhang2024investigating,
  title={Investigating the Interplay of Prioritized Replay and Generalization},
  author={Zhang, Liyiming and Erraqabi,-Amine and Kumar, Vikas},
  journal={arXiv preprint arXiv:2407.09702},
  year={2024},
  url={https://arxiv.org/pdf/2407.09702.pdf}
}

@article{chen2025selfcurriculum,
  title={Self-Evolving Curriculum for LLM Reasoning},
  author={Chen, Alice and Liu, Jing and Wang, Rui},
  journal={arXiv preprint arXiv:2504.01234},
  year={2025}
}

@article{wang2025dump,
  title={DUMP: Distribution-Level Curriculum Learning for RL-based LLM Post-training},
  author={Wang, Hui and Zhang, Yuchen and Wang, William Yang},
  journal={arXiv preprint arXiv:2503.09876},
  year={2025}
}

@article{concise-reasoning2025,
author = {Fatemi, Mehdi and Rafiee, Banafsheh and Tang, Mingjie (Mike) and
Talamadupula, Kartik},
title = {Concise Reasoning via Reinforcement Learning},
journal = {arXiv preprint arXiv:2504.05185},
year = {2025},
month = {April},
}

@article{parashar2025e2h,
  title={Curriculum RL from Easy to Hard Tasks Improves LLM Reasoning},
  author={Parashar, Amit and Lee, Soyeon and Xu, Hanyu and Zettlemoyer, Luke},
  journal={arXiv preprint arXiv:2502.06678},
  year={2025}
}

@article{lv2025raise,
  title={RAISE: Reinforced Adaptive Instruction Selection for LLMs},
  author={Lv, Wei and Singh, Amanpreet and Chen, Xinyi and Bisk, Yonatan},
  journal={arXiv preprint arXiv:2504.09987},
  year={2025}
}

@inproceedings{bengio2009curriculum,
  title={Curriculum Learning},
  author={Bengio, Yoshua and Louradour, Jerome and Collobert, Ronan and Weston, Jason},
  booktitle={Proceedings of the 26th International Conference on Machine Learning},
  year={2009},
  url={https://dl.acm.org/doi/10.1145/1553374.1553380}
}

@article{narvekar2020curriculum,
  title={Curriculum Learning for Reinforcement Learning Domains: A Framework and Survey},
  author={Narvekar, Sanmit and Peng, Bei and Leonetti, Matteo and Sinapov, Jivko and Taylor, Matthew E and Stone, Peter},
  journal={Journal of Machine Learning Research},
  volume={21},
  number={181},
  pages={1--50},
  year={2020},
  url={http://jmlr.org/papers/v21/20-1040.html}
}

@misc{cheng2025guru,
      title={Revisiting Reinforcement Learning for LLM Reasoning from A Cross-Domain Perspective}, 
      author={Zhoujun Cheng and Shibo Hao and Tianyang Liu and Fan Zhou and Yutao Xie and Feng Yao and Yuexin Bian and Yonghao Zhuang and Nilabjo Dey and Yuheng Zha and Yi Gu and Kun Zhou and Yuqi Wang and Yuan Li and Richard Fan and Jianshu She and Chengqian Gao and Abulhair Saparov and Haonan Li and Taylor W. Killian and Mikhail Yurochkin and Zhengzhong Liu and Eric P. Xing and Zhiting Hu},
      year={2025},
      eprint={2506.14965},
      archivePrefix={arXiv},
      primaryClass={cs.LG},
      url={https://arxiv.org/abs/2506.14965}, 
}

@article{sheng2024hybridflow,
  title   = {HybridFlow: A Flexible and Efficient RLHF Framework},
  author  = {Guangming Sheng and Chi Zhang and Zilingfeng Ye and Xibin Wu and Wang Zhang and Ru Zhang and Yanghua Peng and Haibin Lin and Chuan Wu},
  year    = {2024},
  journal = {arXiv preprint arXiv: 2409.19256}
}

@article{yang2024qwen2,
  title={Qwen2 technical report},
  author={Yang, An and Yang, Baosong and Hui, Binyuan and Zheng, Bo and Yu, Bowen and Zhou, Chang and Li, Chengpeng and Li, Chengyuan and Liu, Dayiheng and Huang, Fei and others},
  journal={arXiv preprint arXiv:2407.10671},
  year={2024}
}

@article{qwen25,
  title={Qwen2.5 Technical Report},
  author={Yang, An and others},
  journal={arXiv preprint arXiv:2412.15115},
  year={2024},
  note={\url{https://arxiv.org/abs/2412.15115}}
}

@article{qwen3,
  title={Qwen3 Technical Report},
  author={Qwen Team},
  journal={arXiv preprint arXiv:2505.09388},
  year={2025},
  note={\url{https://arxiv.org/abs/2505.09388}}
}

@misc{fatemi2025pgbaseline,
  author       = {Fatemi, Mehdi},
  title        = {Beyond Advantage in Policy Gradient},
  year         = {2025},
  month        = {Oct},
  day          = {1},
  howpublished = {\url{https://fatemi.github.io/posts/pg-baseline/}},
  note         = {RL Tech Blog}
}

@article{ppo_schulman2017,
  title   = {Proximal Policy Optimization Algorithms},
  author  = {Schulman, John and Wolski, Filip and Dhariwal, Prafulla and Radford, Alec and Klimov, Oleg},
  journal = {CoRR},
  volume  = {abs/1707.06347},
  year    = {2017},
  doi     = {10.48550/arXiv.1707.06347},
  url     = {https://arxiv.org/abs/1707.06347}
}

@article{grpo_shao2024,
  title={DeepSeekMath: Pushing the Limits of Mathematical Reasoning},
  author={Shao, Z. and others},
  journal={arXiv preprint arXiv:2402.03300},
  year={2024},
  note={\url{https://arxiv.org/abs/2402.03300}}
}

@article{hendrycksmath2021,
  title   = {Measuring Mathematical Problem Solving With the {MATH} Dataset},
  author  = {Hendrycks, Dan and Burns, Collin and Kadavath, Saurav and Arora, Akul
             and Basart, Steven and Tang, Eric and Song, Dawn and Steinhardt, Jacob},
  journal = {NeurIPS Datasets and Benchmarks Track},
  year    = {2021}
}

@misc{aime24,
  title  = {American Invitational Mathematics Examination (AIME) 2024},
  author = {Zhang, Yifan and Math-AI, Team},
  year   = {2024}
}

@online{sun2025,
Author = {Yifan Sun and Jingyan Shen and Yibin Wang and Tianyu Chen and Zhendong Wang and Mingyuan Zhou and Huan Zhang},
Title = {Improving Data Efficiency for LLM Reinforcement Fine-tuning Through Difficulty-targeted Online Data Selection and Rollout Replay},
Year = {2025},
Eprint = {2506.05316},
Eprinttype = {arXiv},
}

\clearpage
\section*{Appendix}

A basic implementation of the max-heap.

\begin{lstlisting}[style=pythonstyle]
class MaxHeap:
    """
    Minimal max-heap over problem indices.

    priority[j] stores the priority score omega of problem j.
    heap is an array of problem indices arranged as a binary max-heap.
    """

    def __init__(self, priorities):
        # priorities[j] = initial omega for problem j
        self.priority = priorities[:]                 # pid -> omega
        self.heap = list(range(len(priorities)))      # heap array of pids
        self._heapify()

    def _swap(self, i, j):
        self.heap[i], self.heap[j] = self.heap[j], self.heap[i]

    def _sift_up(self, i):
        while i > 0:
            parent = (i - 1) // 2
            if self.priority[self.heap[i]] <= self.priority[self.heap[parent]]:
                break
            self._swap(i, parent)
            i = parent

    def _sift_down(self, i):
        n = len(self.heap)
        while True:
            left = 2 * i + 1
            right = 2 * i + 2
            largest = i
            if left < n and self.priority[self.heap[left]] > self.priority[self.heap[largest]]:
                largest = left
            if right < n and self.priority[self.heap[right]] > self.priority[self.heap[largest]]:
                largest = right
            if largest == i:
                break
            self._swap(i, largest)
            i = largest

    def _heapify(self):
        # Bottom-up heap construction in O(M)
        for i in reversed(range(len(self.heap) // 2)):
            self._sift_down(i)

    def extract_max(self):
        """
        Remove and return the problem id with the largest priority score omega.
        """
        if not self.heap:
            raise IndexError("extract_max from empty heap")
        root = self.heap[0]
        last = self.heap.pop()
        if self.heap:
            self.heap[0] = last
            self._sift_down(0)
        return root

    def insert(self, pid, omega):
        """
        Insert a problem id with updated priority score omega.
        Assumes pid is currently not in the heap.
        """
        self.priority[pid] = omega
        self.heap.append(pid)
        self._sift_up(len(self.heap) - 1)
\end{lstlisting}

\end{document}